\documentclass[twocolumn,letterpaper]{IEEEAerospaceCLS}  

\newcommand{\Abf}{\boldsymbol{A}}

\newcommand{\Bbf}{\boldsymbol{B}}

\newcommand{\pbf}{\boldsymbol{p}}

\newcommand{\sbf}{\boldsymbol{s}}
\newcommand{\Sbf}{\boldsymbol{S}}

\newcommand{\Scal}{\mathcal{S}}

\newcommand{\Tcal}{\mathcal{T}}

\newcommand{\ubf}{\boldsymbol{u}}

\newcommand{\Vcal}{\mathcal{V}}

\newcommand{\deltabf}{\boldsymbol{\delta}}

\usepackage{amsmath}
\usepackage{amssymb}
\usepackage{amsthm}
\usepackage{amsfonts}

\usepackage[]{graphicx}    
\newcommand{\ignore}[1]{}  
\begin{document}
\title{Machine Learning Based Relative Orbit Transfer for Swarm Spacecraft Motion Planning}

\author{%
Alex Sabol\\ 
The University of Texas at Arlington\\
701 S. Nedderman Dr. \\
Arlington, TX 76019 \\
jonathan.sabol@mavs.uta.edu
\and 
Kyongsik Yun\\
Jet Propulsion Laboratory\\
California Institute of Technology\\ 
4800 Oak Grove Dr., Pasadena, CA 91109 \\
kyongsik.yun@jpl.nasa.gov
\and
Muhammad Adil\\ 
The University of Texas at Arlington\\
701 S. Nedderman Dr. \\
Arlington, TX 76019 \\
muhammad.adil@mavs.uta.edu
\and 
Changrak Choi\\
Jet Propulsion Laboratory\\
California Institute of Technology\\ 
4800 Oak Grove Dr., Pasadena, CA 91109 \\
changrak.choi@jpl.nasa.gov
\and
Ramtin Madani\\ 
The University of Texas at Arlington\\
701 S. Nedderman Dr. \\
Arlington, TX 76019 \\
ramtin.madani@uta.edu
\thanks{\footnotesize 978-1-6654-3760-8/22/$\$31.00$ \copyright2022 IEEE}              
}

\maketitle

\thispagestyle{plain}
\pagestyle{plain}

\maketitle

\thispagestyle{plain}
\pagestyle{plain}
\begin{abstract}
%
In this paper we describe a machine learning based framework for spacecraft swarm trajectory planning. In particular, we focus on coordinating motions of multi-spacecraft in formation flying through passive relative orbit(PRO) transfers. Accounting for spacecraft dynamics while avoiding collisions between the agents makes spacecraft swarm trajectory planning difficult. Centralized approaches can be used to solve this problem, but are computationally demanding and scale poorly with the number of agents in the swarm. As a result, centralized algorithms are ill-suited for real time trajectory planning on board small spacecraft (e.g. CubeSats) comprising the swarm. In our approach a neural network is used to approximate solutions of a centralized method. The necessary training data is generated using a centralized convex optimization framework through which several instances of the n=10 spacecraft swarm trajectory planning problem are solved. We are interested in answering the following questions which will give insight on the potential utility of deep learning-based approaches to the multi-spacecraft motion planning problem: 1) Can neural networks produce feasible trajectories that satisfy safety constraints (e.g. collision avoidance) and low in fuel cost? 2) Can a neural network trained using n spacecraft data be used to solve problems for spacecraft swarms of differing size?
\end{abstract} 

\tableofcontents

\section{Introduction}
Multi-agent systems have shown great flexibility and robustness, achieving impressive results across a wide variety of application domains \cite{rossi2018review}. For example in \cite{vinyals2019grandmaster} multi-agent reinforcement learning was used to achieve near-professional levels of play in the real time strategy game \textit{Starcraft II}. A multi-agent deep-RL approach is used to control traffic signals in \cite{chu2019multi}. The authors of \cite{bu2019smart} utilize a multi-agent system to improve agricultural production. Despite a growing body of literature demonstrating the success of multi-agent systems, their flexibility comes at a price, as the coordination of multi-agent systems can be computationally expensive \cite{demaine2018toward}. Recent advances in theory coupled with growing computational power has made multi-agent problems more tractable, resulting in increased interest in multi-agent applications to the spacecraft domain. In this paper we investigate spacecraft swarm trajectory planning in the context of passive relative orbit (PRO) transfers. 
\par
Spacecraft swarms are potentially a transformational element for future space exploration and science missions. The low-cost, versatility through transformation, and robustness to failure that a swarm of small spacecraft can provide has great benefit over the current monolithic counterpart. For such missions, autonomy is crucial in its operation, and how to plan coordinated motions of multi-spacecraft plays a key role. Spacecraft swarms have been well studied in the literature, see \cite{teja2019attitude},\cite{nallapu2019spacecraft},\cite{dono2018propulsion},\cite{koenig2018safe} for examples of recent work in spacecraft swarms. For a survey on formation flying see \cite{di2018survey}. \par
Spacecraft swarm trajectory planning is related to a more general problem in robotics, called the \textit{multi-robot path planning(MRPP)} or the \textit{multi-agent path finding(MAPF)} problem. As with swarm trajectory planning,  collision-free paths must be generated, which is well-known to be NP-hard \cite{johnson2018relationship}. Many works such as \cite{atias2018effective},\cite{sivanathan2020decentralized},\cite{luis2020online},\cite{wang2021research} have explored multi-robot path planning in more general contexts.
\par In a sequential approach to spacecraft swarm trajectory planning, the goal position at the next time step is calculated at each time step for each spacecraft in the swarm. Alternatively, the trajectories can be planned from start to finish over the full time horizon. These approaches can be further categorized as either \textit{centralized}, or \textit{decentralized} approaches.


\par In \textit{centralized} approaches, the path for each spacecraft is computed by a central authority and then communicated back to the spacecraft.
Sampling-based algorithms \cite{shome2020drrt}, \cite{le2019multi} have been some of the most successful methods for solving the centralized problem, as they are proven to achieve asymptotic optimality. In \cite{sharon2015conflict}, the conflict-based search algorithm is able to solve the  centralized multi-robot path planning problem to optimality. Centralized approaches can generate globally optimal solutions to the spacecraft swarm trajectory planning problem, but have low convergence rates and do not scale well with the number of spacecraft. For real-time applications, centralized methods may not be suitable for two reasons: First, solving the problem can take the centralized authority too long, making real-time applications infeasible, and second, there are many situations in which communication between the spacecraft and a central authority is impractical or even impossible.
\par On the other hand in \textit{decentralized} approaches to spacecraft swarm planning, each spacecraft calculates its own trajectory using local information. These approaches sacrifice the solution quality of centralized approaches in favor of computational speed and tractability, allowing for real-time applications on modest hardware in some cases. For example the distributed algorithm presented in \cite{bandyopadhyay2017dist} runs on-board small spacecraft, and is able to handle instances of the spacecraft swarm trajectory planning problem with up to 50 unique spacecraft. In \cite{honig2018trajectory}, a decentralized approach to trajectory planning for quadrotor swarms is presented. Decentralized approaches commonly use sequential algorithms, as planning over the full time horizon can be computationally prohibitive. \par 
Swarm trajectory planning can be formulated as a convex programming problem, allowing researchers to take advantage of many sophisticated theoretical results and algorithms \cite{boyd2004convex}. For example, in \cite{morgan2014model} sequential convex programming is used to develop a decentralized trajectory planning algorithm for spacecraft swarm control. Additionally, sequential convex programming is used for distributed motion planning in cluttered environments in \cite{bandyopadhyay2017distributed}, and the authors of \cite{goel2017trajectory} utilize sequential convex programming for trajectory planning applications to space-based solar power. For additional applications of convex programming to spacecraft swarm formation flying see \cite{chen2015decoupled},\cite{wang2017minimum},\cite{morgan2016swarm},\cite{misra2017optimal},\cite{foust2016autonomous}. \par
In addition to conventional algorithms, machine learning-based approaches have also been applied to trajectory planning problems. For example, in \cite{li2020graph}, a convolutional neural network and a graph neural network are used to solve the decentralized problem. In this case, the authors used conflict-based search to generate their supervised learning data.
It is difficult to utilize supervised learning schemes for spacecraft swarm trajectory planning, as the centralized problem must be solved many times to generate the necessary training data. As a result, reinforcement learning has been more commonly applied to MRPP, resulting in a large body of literature. In \cite{lin2020dynamic}, deep reinforcement learning is applied to UAV formation flying. The authors of \cite{zhao2020robust} use deep RL to coordinate the movement of cooperating quadrotors. In \cite{sartoretti2019primal}, a reinforcement learning framework is used to solve the multi-agent path planning problem. Their framework is not fully unsupervised - the networks were also augmented with example solutions produced using an expert algorithm. Many other examples of reinforcement learning applications for spacecraft swarm formation flying can be seen in the literature, e.g.  \cite{liu2019attitude}, \cite{zhang2020lyapunov}, \cite{liu2020reinforcement}, \cite{sui2020formation}.
\par A key idea motivating the work in this paper is: If we were able to solve the full trajectory planning problem in real time on-board each robot, then for each robot only the initial and goal states of the other robots need to be communicated.
We present a supervised learning framework in which we utilize training data generated offline by a convex optimization solver. We take this approach because the computational burden comes from generating the training data and training the neural network, both of which can be done offline in advance. The resulting trained neural network requires little computational power and can be uploaded to each spacecraft. Furthermore, the neural networks imitate a centralized trajectory planner, which generates a complete set of trajectories over the full time horizon. This is in contrast to many online decentralized approaches where a sequential scheme is used by each spacecraft - at each time step the spacecraft calculate their position at the next time step. One reason that a trajectory planning approach over the complete time horizon is useful is because the sequential scheme relies on spacecraft communicating with each other in between each time step. \par Since generating the training data becomes increasingly prohibitive as the number of spacecraft is increased, we are interested in our neural networks ability to generalize. In particular, we investigate whether our neural network trained on n spacecraft problems can be used to solve problems with more than n spacecraft. Additionally, we test the neural networks' ability to generalize to problems with different transfer times. The transfer time is the amount of time available for each spacecraft to travel from its initial state to its goal state. Transfer time is particularly interesting in our context, as the optimal trajectory shape is dependent on the transfer time. For example, shorter transfer times tend to produce paths which are closer to straight lines, and longer transfer times lead to paths with more curvature.  \par
Our contributions can be summarized as follows: \begin{itemize}
\item The development of a supervised learning based approach to the spacecraft swarm trajectory planning problem which is fast enough for real-time applications.
\item The comparison of several neural network architectures with respect to their ability to generalize to larger problem sizes - we compare traditional feed forward neural network with LSTM neural networks, analyzing solution quality in terms of the number of collisions, and fuel optimality.
\item We integrate our machine learning approach with a conventional convex optimization-based approach to the swarm trajectory planning problem.. By using our neural network to produce the initial seed, the convex optimizer is able to converge quicker when solving the swarm trajectory planning problem.
\end{itemize}
\subsection{Paper Organization}
The remainder of the paper is organized as follows. In Section \ref{problem}, the spacecraft swarm trajectory planning problem in the context of PRO transfer is described in more detail. In Section \ref{learning} we briefly review neural network architectures, and describe our supervised learning approach to the swarm trajectory planning problem. In Section \ref{results} we discuss our experimental results. Finally, Section \ref{conclusion} concludes the paper, and possible directions for future work are suggested.

\section{Problem Description} \label{problem}

\begin{figure*}[t] 
    \centering
	\includegraphics[width=1.0\linewidth]{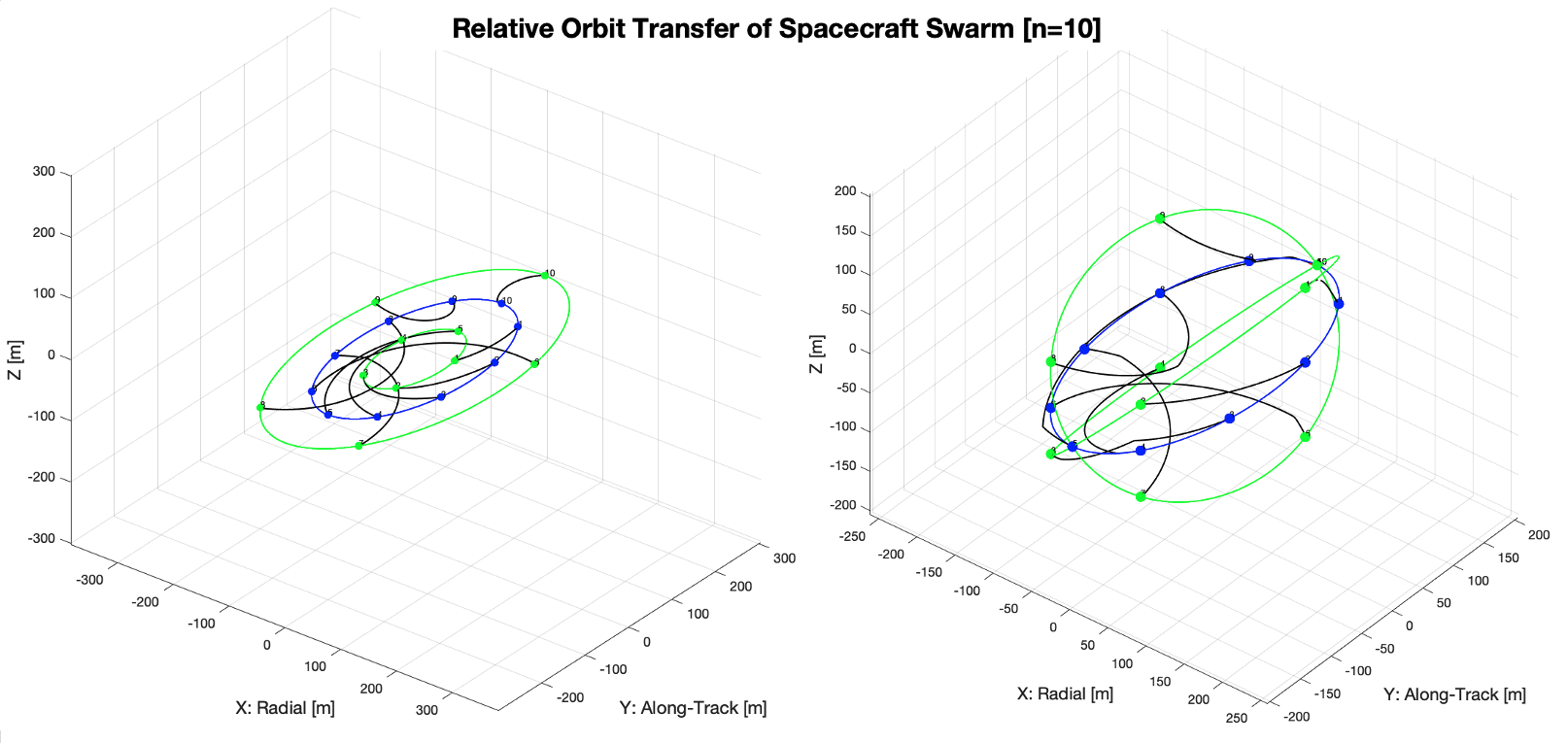}
	\caption{\textbf{Examples of relative orbit transfer trajectories for a fleet of spacecraft are shown in LVLH frame for two problem instances. The initial states of spacecraft on current relative orbits are shown in blue, and goal states of spacecraft on goal relative orbits are shown in green. In both examples, 10 spacecraft are initially equally spaced out in one relative orbit with semi-major axis of 200m. On the left, the goal states require that the spacecraft transfer to two relative orbit in the same plane, five spacecraft on each, with semi-major axis of 100m and 300m respectively. On the right, spacecraft are transferred to two relative orbits, five on each, that are slanted $\pm$45 deg from the initial relative orbit. The minimum fuel trajectories that transfer spacecraft between the given relative orbits are shown in black.} }
	\label{fig:rel_orbit_transfer_example}
\end{figure*}

Our focus is on solving swarm spacecraft motion planning problems that transfer a fleet of spacecraft from one set of relative orbits to another. Spacecraft within the fleet are assumed to be formation flying in close proximity to each other. The initial and goal relative orbits are assumed to satisfy the energy matching condition such that the relative motions are bounded. 

Let $\Vcal=\{$$v_1,...,v_n$$\}$ be a set of \textit{n} spacecraft and $\Tcal=\{$$1,2,...,T$$\}$ be the discrete time horizon of the trajectory planning. Let $\sbf_i[t] = [x_i ~ y_i ~ z_i ~ \dot x_i ~ \dot y_i ~ \dot z_i]^{\top}$ be state, both position and velocity, of a spacecraft $i$ at time $t \in \Tcal$. Similarly, let $\pbf_i[t] = [x_i ~ y_i ~ z_i]^{\top}$ be position of a spacecraft $i$ at time $t$. In the problem, we are given a set of initial and goal states, ${\Scal}^{\mathrm{init}}=\{\sbf_i[1]\}_{i=1}^n$, ${\Scal}^{\mathrm{goal}}=\{\sbf_i[T]\}_{i=1}^n$ that represents states of each spacecraft in their current and goal relative orbits, respectively, at the start and end of the transfer. The task is to generate a set of transfer trajectories that are dynamically feasible, safe, and optimal with respect to a given objective cost.

\subsection{Spacecraft Dynamics}
We consider a model in which \textit{n} spacecraft are formation flying in elliptical orbits around the Earth. With the assumption that the swarm of spacecraft are in close proximity to each other, we will utilize the well-known Clohessy–Wiltshire(CW) equations to represent dynamics of the relative motions. The CW equations are a simplified model of orbital relative dynamics linearized about the moving circular reference frame of the Earth \cite{curtis2013orbital}. In the CW model, each spacecraft's dynamics can be expressed as
\begin{align} \label{CW}
    \ddot x&=  ~3e^2x + 2e\dot y + u_1  \\
    \ddot y&=  -2e\dot x + u_2 \nonumber\\
    \ddot z&=  -e^2z + u_3 \nonumber\\
         e &= \sqrt{\frac{\mu}{r_0^3}} \nonumber
\end{align}
in the Local-Vertical, Local-Horizontal (LVLH) coordinate system centered at the target circular orbit. The x-axis points radially away from the Earth, y-axis is in the direction of the orbital velocity (along-track) of the target, and z-axis completes the right handed system that also corresponds to direction of the angular momentum of the target. The parameter $e$ is the mean motion of the target, where $\mu$ is the Earth's gravitational constant, and $r_0$ is the radius of the target orbit. Spacecraft control input vector is represented by $\ubf = [u_1 ~ u_2 ~ u_3]$.

\par
The CW equations have a closed form solution \cite{wh1960terminal} in continuous time, and discrete time dynamics can be derived from it. The discrete dynamics of each spacecraft can be written as
\begin{equation}
\sbf_i[t+1] = \Abf\sbf_i[t]+\Bbf\ubf_i[t]    \quad\quad \forall t\! \in\! \Tcal
\end{equation}
where $\Abf \in {\mathbb{R}}^{6\times6}, \Bbf \in {\mathbb{R}}^{6\times3}$ are system matrices obtained by solving and discretizing the CW equations. We utilize this discrete dynamics when planning the transfer trajectories over a given discrete time planning horizon.

\subsection{Relative Orbit Transfer Trajectories}

Each spacecraft $i$ must transfer from its initial state $\sbf_i[0]$ on the current relative orbit at $t=0$, to the goal state $\sbf_i[T]$ on a different relative orbit at $t=T$. The initial and goal states are assumed to satisfy energy matching condition
\begin{equation}
\dot y(0) = -2e x(0), ~~ \dot y(T) = -2ex(T)
\end{equation}
such that the start and end relative orbits are bounded. The relative orbits under consideration can be viewed as \emph{passive relative orbits} (PROs) that are thrust-free periodic relative spacecraft trajectories in the given LVLH frame.

The transfer trajectory of spacecraft $i$ in discrete time can be represented as the matrix $\Sbf_i = [\sbf_i[1] ~ \sbf_i[2] ~ ... ~ \sbf_i[$T$]] \in {\mathbb{R}}^{6\times T}$, where the vector $\sbf_i[t]$ represent state of spacecraft $i$ at each time-step as previously defined. The transfer trajectories can be optimized with respect to a customized cost that reflects the mission objective of the spacecraft swarm. Here, we choose a general objective of minimizing fuel cost of the transfer that is often of concern for small spacecraft with limited delta-V budget.



\subsection{Fuel Optimality}
For each spacecraft, the fuel consumption is proportional to the control input $\ubf$, which makes minimizing the sum of $L_1$-norms of the control inputs for all spacecraft across our time horizon a straightforward objective.
\begin{equation}
    f_{fuel} = \sum_{t=1}^T\sum_{i=1}^n || \ubf_i[t] ||_1
\end{equation}
Note that the objective is to minimize the total fuel expenditure of all $n$ spacecraft involved in the transfer. This requires concurrently planning all transfer trajectories in the joint state-space of $n$ spacecraft.

\subsection{Collision Avoidance}
The most important aspect of transfer trajectory planning is avoiding collisions between the spacecraft as they are formation flying in close proximity. We model each spacecraft as a sphere with radius $r$ and require that spacecraft keep a buffer distance of $r_b$ between each other. We define the collision radius as $r_{col} = 2r + r_b$, and if the distance between spacecraft $i$ and $j$ are within this radius at time t, then we consider two spacecraft to have collided. We would like our trajectory set to contain a minimal number of collisions. The total number of collisions, $f_{col}$, can be defined using the following equations, where $\pbf_i[t]$, $\pbf_j[t]$ are positions of spacecraft $i$ and $j$ at time $t$ as previously defined.
\[ \deltabf(\pbf_{i}[t],\pbf_{j}[t])=
\begin{cases} 
      1, & ||\pbf_i[t] - \pbf_j[t] ||_2\leq r_{col} \\
      0, & ||\pbf_i[t] - \pbf_j[t] ||_2> r_{col} 
   \end{cases}\]
\begin{equation}
    f_{col} = \sum_{t=1}^T\sum_{i=1}^{n-1}\sum_{j=i+1}^n \deltabf(\pbf_{i}[t],\pbf_{j}[t])
\end{equation}
In this paper our goal is to learn a mapping from the initial and goal states $({\Scal}^{\mathrm{init}},{\Scal}^{\mathrm{goal}})$, to a set of trajectories, $\Sbf^*=\{\Sbf_1,\Sbf_2,...,\Sbf_n\}$, which is optimal with respect to fuel consumption and satisfy collision avoidance. Given the input states, our trained architecture should produce an estimate of the centralized trajectory plan that would be obtained by our expert algorithm, which is a follow-up of the sequential convex programming framework presented in \cite{morgan2016swarm}. 

\section{Supervised Learning Approach} \label{learning}
Our feed forward neural networks consist of several densely connected layers with an activation function and a dropout layer \cite{yun2020multi}. We used rectified linear units for the nonlinear activation function: $g(x)=\mathrm{max}(0,x)$ where $x$ is the input to the neuron. Rectifier is the most used activation function for deep neural networks with better gradient propagation that has neuro-computational plausibility and less vanishing gradient problems. Dropout regularization rate of 0.5 was used to prevent overfitting. 

Neural network training aims to minimize the error function $E$, which is the mean squared error (MSE) that quantifies the difference between the computed output trajectory (feed-forward neural network) and the actual trajectory (mathematical model). The number of layers and parameters were determined empirically by examining the performance of several different layers (3-12) and parameters (10-200). We found that 100 parameters per layer and 4 total layers gave the least mean squared error ($MSE<0.001$).

The training data we used was generated using a sequential convex optimization-based framework which has been developed by us and will be the focus of a future publication. The convex framework used to generate our training data is discussed in \cite{choi2021optimal}. Although this framework is too computationally demanding to be used by the spacecraft directly, it is tractable enough that it can be used to generate training data for the neural networks.


A disadvantage of traditional feed forward neural network architectures is the need to explicitly define the input data size. For a single instance of the spacecraft swarm trajectory planning problem, our input and output data consists of n spacecraft trajectories concatenated horizontally. Each trajectory consists of the x,y, and z positions and velocities for each discrete time step 1 through T. Thus the input for each problem instance is a 6 x nT matrix. Since our neural networks were trained on n=10 and T=11 data, the input size was fixed as a 6 x 110 matrix. Testing trajectory planning problems involving less than ten spacecraft is straightforward, as an equivalent ten spacecraft problem can be produced by padding the input with zeros. In other words, solving the trajectory planning problem for four spacecraft is equivalent to solving a ten spacecraft trajectory planning problem with the four original spacecraft and six 'virtual' spacecraft which remain stationary. For problems involving more than ten spacecraft, constructing an equivalent input for our feed forward neural networks was nontrivial. For n spacecraft problems, we used the feed-forward neural networks to solve corresponding ten spacecraft problems for many different combinations of the n agents. The trajectories produced while solving these ten spacecraft problems are then averaged to generate the trajectories for the full n agent problem. Note that for a given n, there are $n\choose10$ possible combinations of ten spacecraft which can be selected from the n spacecraft. Testing every ten agent combination was impractical for large n, so the number of ten agent combinations tested was limited to 100 per problem instance. For reference, in a single instance of the spacecraft trajectory planning problem for n=30, there are 30,045,015 possible combinations of 10 unique spacecraft.

A Long short-term memory (LSTM) neural network is a recurrent neural network(RNN) architecture which is well-suited for problems involving time-series data, such as trajectory planning. \cite{gers2000learning} An LSTM unit is commonly made up of an input gate, a cell, a forget gate, and an output gate. LSTM architectures are designed to deal with the vanishing gradient problem during backpropagation, which prevents other RNN architectures from learning long-term dependencies in the data \cite{calin2020deep}. LSTM architectures were convenient for our purposes as they do not require the input data size to be explicitly defined. This allowed us to easily use data for any number of spacecraft directly as the input to our LSTMs, which were trained on n=10 agent data. This is in contrast to traditional feed-forward neural networks, which require an extensive amount of preprocessing and postprocessing if the input problem has a different number of spacecraft than the neural network was trained on.

One drawback that can be encountered when solving physical problems using supervised learning schemes is that neural networks are designed to simply learn mappings from input data to output data, and do not explicitly incorporate the problem's physical constraints into the reward function. This fact has not gone unnoticed by researchers, and references to 'physics-informed neural networks can be found in the literature, e.g. \cite{pagnier2021physics}, \cite{karniadakis2021physics}. In this work we have briefly investigated physics informed neural networks in the context of spacecraft swarm trajectory planning. In two of the neural network models tested(FF$_{col}$ and LSTM$_{col}$) the regression output layer has been replaced by a custom output layer which explicitly penalizes spacecraft collisions. 
\section{Results} \label{results}
All experiments were done using MATLAB R2020b with the Deep Learning Toolbox for neural network support. We test six different neural network architectures - three feed forward neural networks and three LSTMs.  A summary of the parameters used in these architectures is shown in table \ref{tab:exp1_model_parameters}.
\begin{table}[h] 
    \centering
    	\caption{\textbf{Neural Network Model Parameters} }
	\begin{tabular}{|c c c|} 
		\hline
		Model & Hidden Layers & Neurons per Layer \\
		\hline\hline
		\hline
		FF$_1$ & 1 & 180\\ 
		\hline
		FF$_2$ & 2 & [180,150]\\
		\hline
		FF$_{col}$ & 1 & 180\\
		\hline
		LSTM$_1$ & 1 & 180\\ 
		\hline
		LSTM$_2$ & 2 & [180,150]\\
		\hline
		LSTM$_{col}$ & 1 & 180\\
		\hline
	\end{tabular}
 \label{tab:exp1_model_parameters}
\end{table}
\subsection{Experiment 1}
\begin{figure}[h] 
    \centering
	\includegraphics[scale=.6]{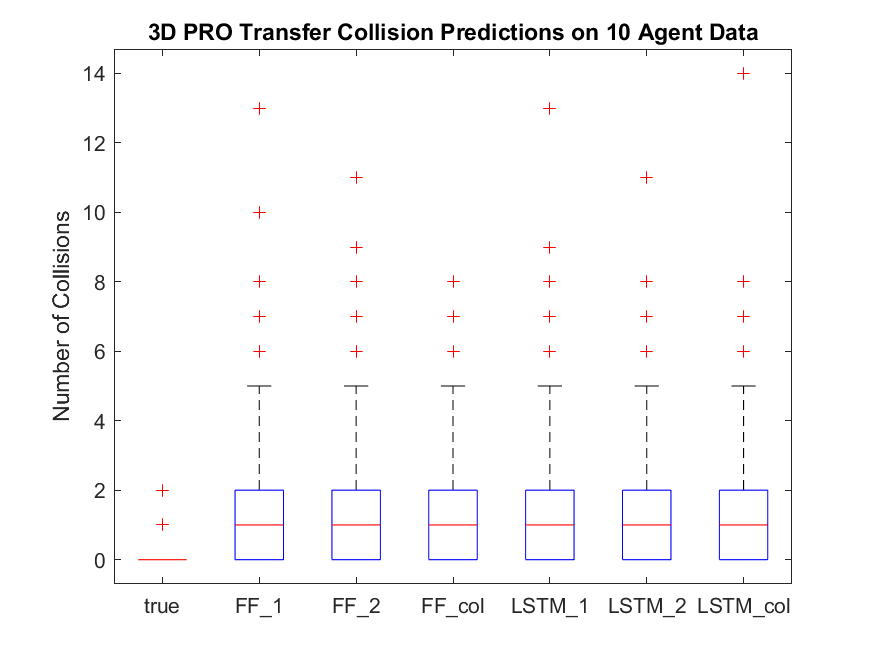}
	\caption{\textbf{Average number of collisions per trial for each neural network architecture. We use 'true' as the label for our ground truth solution produced by the convex optimizer.} } \label{fig:exp1_col}
\end{figure}

In the first experiment we test our framework's ability to solve the spacecraft swarm trajectory planning problem. We test three feed-forward neural network architectures and three LSTM architectures. In the LSTM architectures a hidden layer contains an LSTM layer followed by a dropout layer, and the neurons per layer column corresponds to the number of neurons in the LSTM layers.s For each neural network, training is done using $N_{train}$=8000 instances of the $n=10$ spacecraft case. A collision distance of $r_{col}=15$ and a constant transfer time, resulting in T=11 time steps. We set aside $N_{valid}$=1000 trials for validation, then evaluate our networks using $N_{test}$=1000 test trials.
The total number of collisions achieved by each network as well as the ground truth produced by the convex optimizer can be seen in figure \ref{fig:exp1_col}.\par All six architectures performed well with respect to the total number of collisions, all achieving less than 2 collisions on average. For reference, the convex optimizer achieved an average of .113 collisions per trial. The maximum number of collisions in a single trial was 13,11,8,13,11, and 14, for FF$_1$,FF$_2$,FF$_{col}$, LSTM$_1$,LSTM$_2$, and LSTM$_{col}$, respectively. For comparison, the convex optimizer achieved an average of .113 collisions per trial, and the maximum number of collisions produced by the convex optimizer in a single trial was 2.\par
\begin{figure}[h]
    \centering
	\includegraphics[scale=.6]{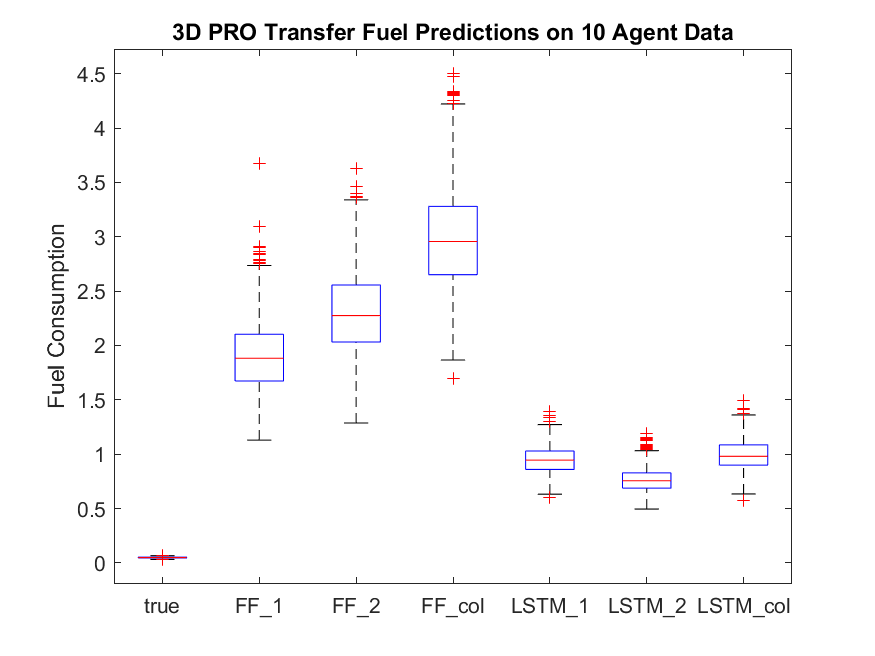}
	\caption{\textbf{Average fuel consumption per trial for each neural network architecture. We use 'true' as the label for our ground truth solution produced by the convex optimizer.} } \label{fig:exp1_fuel}
\end{figure}

The results with respect to fuel consumption can be seen in figure \ref{fig:exp1_fuel}. Here we notice a larger difference between the models; the LSTM architectures produced cheaper trajectories with respect to fuel consumption than the feed-forward neural networks.
A summary of the neural networks' fuel and collision results for the n=10 spacecraft case is shown in table \ref{tab:exp1_obj}, and an example of the trajectories produced by the neural networks is shown in figure \ref{fig:exp1_traj}, which can be found in the appendix.
\begin{table}[h] 
    \centering
    \caption{\textbf{Neural Network Objective Values } }
	\begin{tabular}{|c c c c|} 
		\hline
		Model & Collisions(av.) & Collisions(max.) & Fuel(av.)  \\
		\hline\hline
		true & 0.113 & 2 & .0494 \\
		\hline
		FF$_1$ & 1.326 & 13 & 1.906 \\ 
		\hline
		FF$_2$ & 1.397 & 11 & 2.307 \\
		\hline
		FF$_{col}$ & 1.440 & 8 & 2.98  \\
		\hline
		LSTM$_1$ & 1.594 & 13 & 0.95 \\ 
		\hline
		LSTM$_2$ & 1.428 & 11 & 0.76 \\
		\hline
		LSTM$_{col}$ & 1.520 & 14  & 0.99 \\
		\hline
	\end{tabular}
	 \label{tab:exp1_obj}
\end{table}
\par
One advantage of the feed-forward architectures is a quicker training time when compared to the LSTMs. To reduce the time spent training, we used a 'ValidationPatience' of 5, which prematurely ends the training if the loss increases on the validation set 5 times in a row. Even still, the feed-forward neural networks train quicker than the LSTMs, as shown in table \ref{tab:exp1_train_time}.
\begin{table}[h] 
    \centering
    \caption{\textbf{Neural Network Training Time}}
	\begin{tabular}{|c c|} 
		\hline
		Model & Train Time(s) \\
		\hline\hline
		\hline
		FF$_1$ & 25.133\\ 
		\hline
		FF$_2$ & 30.079\\
		\hline
		FF$_{col}$ & 264.429\\
		\hline
		LSTM$_1$ & 206.810\\ 
		\hline
		LSTM$_2$ & 1291.4\\
		\hline
		LSTM$_{col}$ & 986.14\\
		\hline
	\end{tabular}
	  \label{tab:exp1_train_time}
\end{table}
\subsection{Experiment 2}
\begin{figure}[h] 
    \centering
	\includegraphics[scale=.6]{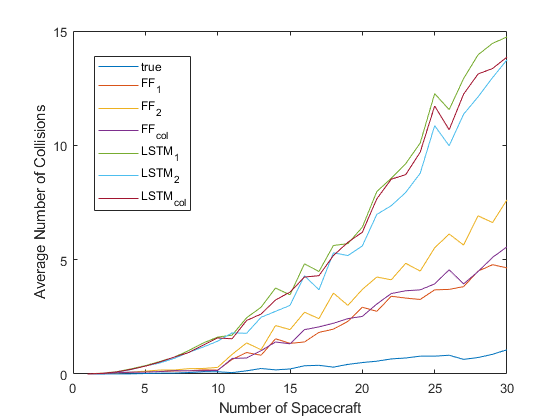}
	\caption{\textbf{Average number of collisions per trial for each neural network architecture for n=2 through n=30 spacecraft problems. We use 'true' as the label for our ground truth solution produced by the convex optimizer.} } \label{fig:exp2_col}
\end{figure}

In the second experiment we tested each model's ability to generalize to varying numbers of spacecraft. We considered swarms of n=2 through n=30 spacecraft for our neural networks which have been trained on exclusively n=10 spacecraft data. As expected, the neural networks produce solutions with an increasing number of collisions as the number of spacecraft is increased, and these solutions are strictly worse than the corresponding solution generated by the convex optimizer. With respect to collisions, FF$_1$ performs the best out of the neural networks, achieving 4.64 collisions per trial on average in the n=30 case. The worst performing neural network is LSTM$_1$, which averages 14.74 collisions per trial in the n=30 case. Surprisingly, all three feed-forward neural networks outperform the LSTMs, achieving less than half of the total number of collisions. These results can be seen in figure \ref{fig:exp2_col}. Although these solutions are promising, the solution produced by the convex optimizer is significantly higher quality, with 1.06 average collisions in the n=30 spacecraft case. 

\begin{figure}[h] 
    \centering
	\includegraphics[scale=.6]{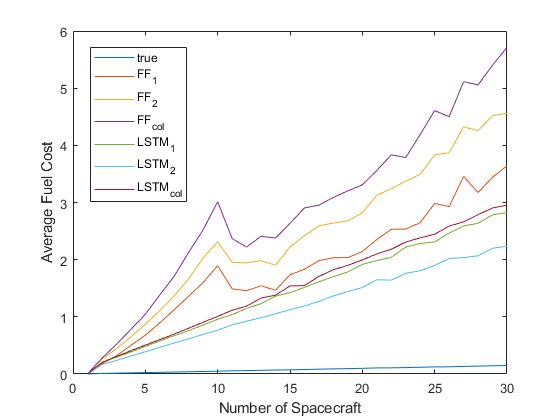}
	\caption{\textbf{Average fuel cost per trial for each neural network architecture for n=2 through n=30 spacecraft problems. We use 'true' as the label for our ground truth solution produced by the convex optimizer.} } \label{fig:exp2_fuel}
\end{figure}

The fuel cost also increases as the number of spacecraft is increased, as shown in figure \ref{fig:exp2_fuel}. LSTM$_2$ performs the best with respect to fuel, obtaining an average fuel consumption of 2.24 on the n=30 spacecraft problem. The worst performing network is $FF_{col}$, which achieves an average fuel consumption of 5.71 on the n=30 spacecraft problem. The LSTMs perform better than the feed forward networks with respect to fuel consumption but a drastic difference in solution quality is not observed. The convex optimizer again outperforms the neural networks, achieving an average fuel consumption of .1484 for the n=30 spacecraft problem.

\begin{figure}[h]
    \centering
	\includegraphics[scale=.6]{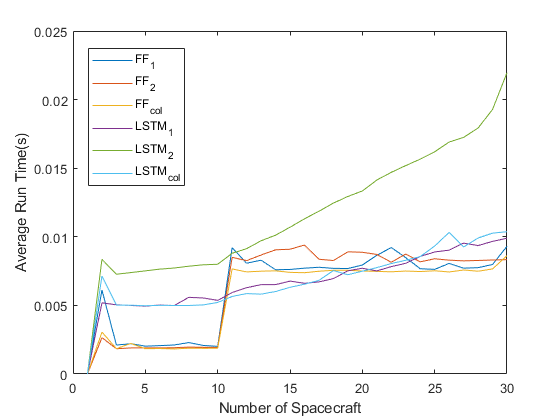}
	\caption{\textbf{Average time taken by the machine learning framework to solve one instance of the trajectory planning problem for n=2 through n=30 spacecraft.} } \label{fig:exp2_runtimes_ml}
\end{figure}

\begin{figure}[h]
    \centering
	\includegraphics[scale=.6]{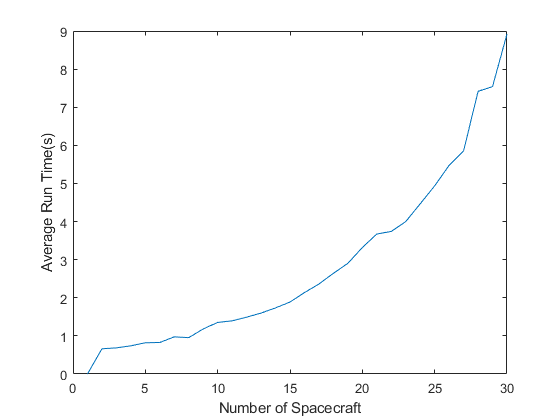}
	\caption{\textbf{Average time taken by the convex optimization framework to solve one instance of the trajectory planning problem for n=2 through n=30 spacecraft.} } \label{fig:exp2_runtimes_opt}
\end{figure}

The average run times for the machine learning framework and the convex optimization framework are reported in figures \ref{fig:exp2_runtimes_ml} and \ref{fig:exp2_runtimes_opt},respectively. For the machine learning frameworks the run time stays relatively constant as the number of spacecraft in the swarm is increased, with the maximum run time taking a few hundredths of a second. In contrast, the convex optimization framework shows a sharp increase in run time as the number of spacecraft is increased, with the n=30 spacecraft problem taking nearly 10 seconds to solve.
\subsection{Experiment 3}
\begin{figure}[h]
	\includegraphics[scale=.22]{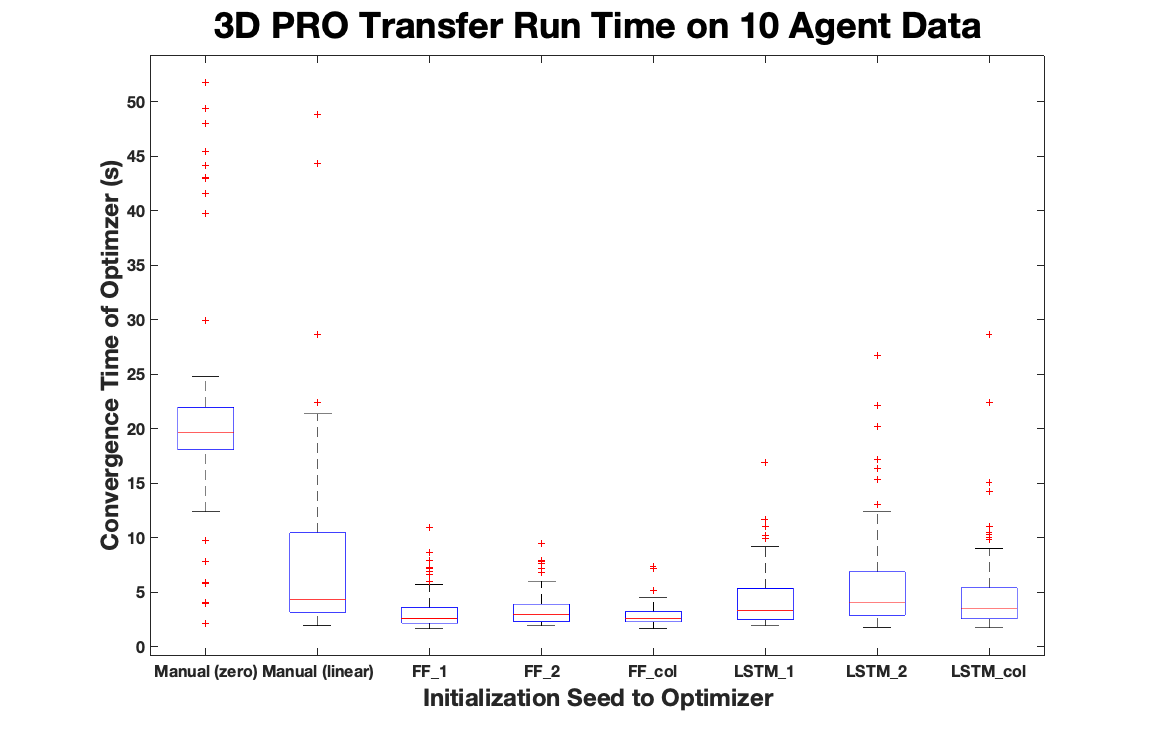}
	\caption{\textbf{Average time taken by the convex optimization framework to solve one instance of the n=10 spacecraft trajectory planning problem. The initialization method used by the convex optimizer is shown on the x axis. The Manual(zero) column refers to the case where no initial seed was used, and in the Manual(linear) column, simple linear interpolation was used as an initial seed. In all other columns, the name of the neural network used to produce the initial seed is reported(e.g. in the column labeled FF$_1$, the convex optimizer used the solution produced by the neural network FF$_1$ as its initial seed).}} \label{fig:exp2_ml_seed}. 
\end{figure}

In the final experiment we test our machine learning framework's ability to interface with the centralized convex optimization planner. We use our machine learning framework to generate initial seeds for the purpose of warm-starting the convex optimization framework. For this experiment, 100 instances of the n=10 spacecraft trajectory planning problem were solved using each of the neural network architectures discussed in table \ref{tab:exp1_model_parameters}. These neural network-produced trajectory plans were then used to initialize the convex optimizer. The average time taken by the convex optimizer to converge to the optimal solution  given differing initial seeds is reported in figure \ref{fig:exp2_ml_seed}. When using the neural network generated seeds the convex optimizer takes less than five seconds on average to converge. In contrast, the convex optimizer takes an average of roughly twenty seconds to converge when no initial seed is provided. Using a simple linear interpolation between the start position and goal position for each spacecraft as an initial seed also reduces convergence time of the convex optimizer on average, but there is a higher variance from trial to trial when compared to neural network seeded initialization. This indicates that seeding by linear interpolation may not reduce the convex optimizer's time to convergence as reliably as a neural network generated initial seed. The speed increase obtained by the convex optimizer when using a neural network generated initial seed did not vary considerably from architecture to architecture, but the quickest convergence was achieved when using FF$_{col}$'s initial seed.


\section{Conclusion} \label{conclusion}
In this work, we presented a machine learning-based approach to the spacecraft swarm trajectory planning problem. We generated solutions with very few collisions and low fuel consumption using relatively simple neural network architectures. Furthermore, we showed that our neural network can be used to initialize a centralized motion planner, reducing its time to convergence. Our preliminary results indicate that generalization to spacecraft swarms of differing size may be possible - despite being trained on the 10 spacecraft case, our neural networks produce promising trajectories for swarms consisting of up to 30 spacecraft. Potential focuses of future work on machine learning based approaches to the spacecraft swarm trajectory planning problem should investigate hyperparameter optimization and collision avoidance with physics informed neural networks. Furthermore, generalization to varying transfer times should be explored in more detail, as the transfer time has a considerable impact on the shape of the trajectories. \\
Despite obtaining preliminary solutions with relatively few collisions, this framework should be expanded upon to ensure collision-free solutions prior to real-world deployment. Nevertheless, our neural network framework can still be utilized to quickly generate initial solutions to be used to warm-start the centralized convex framework seen in \cite{choi2021optimal}.
\acknowledgments
This research was carried out at the Jet Propulsion Laboratory, California Institute of Technology, under a contract with the National Aeronautics and Space Administration. \\ 
\copyright 2021 California Institute of Technology.

\bibliographystyle{IEEEtran}
\bibliography{ref}




\thebiography

\begin{biographywithpic}
{Alex Sabol}{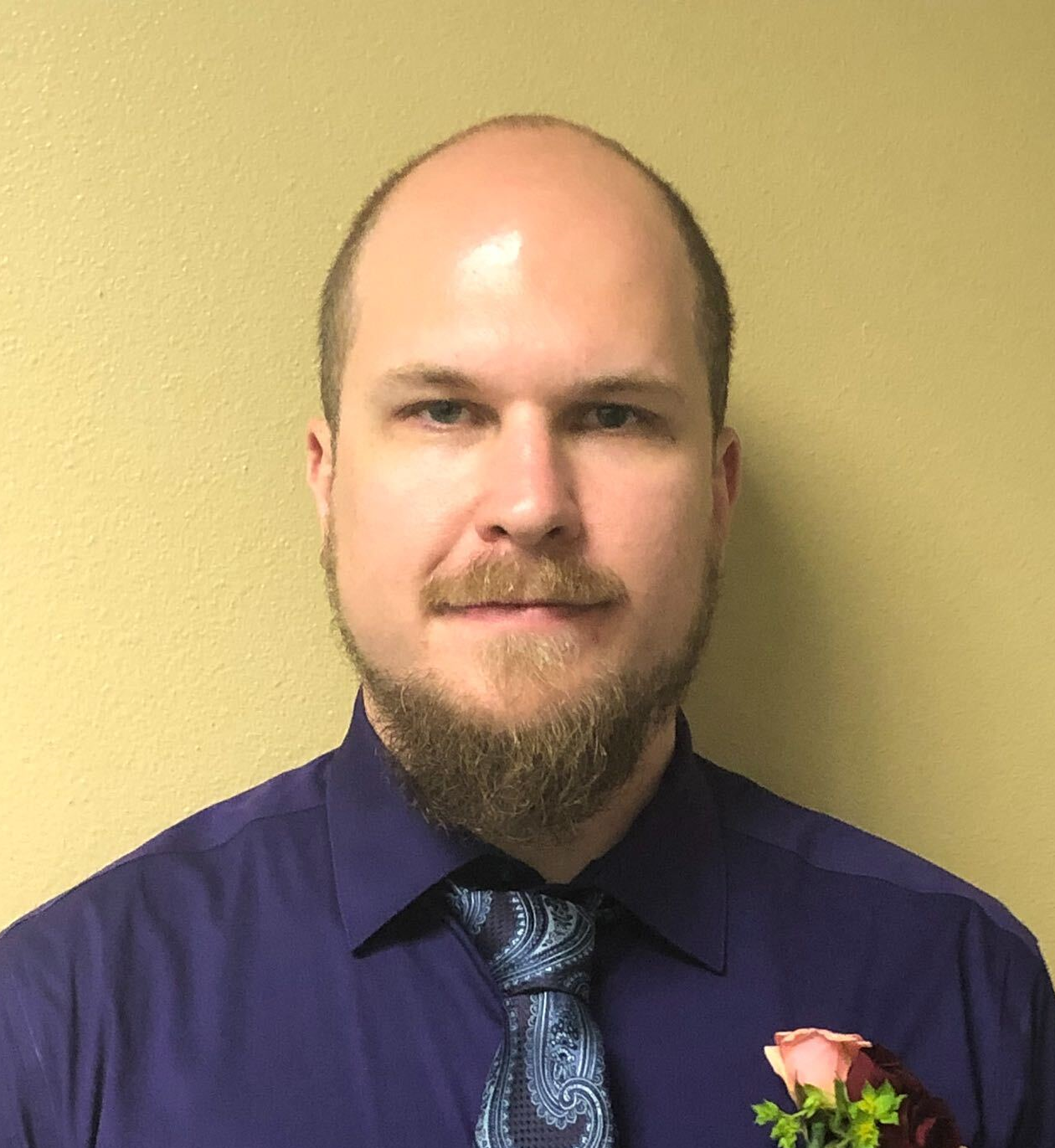}  received Bachelor's Degrees in Mathematics and Chemistry from the University of Texas at Austin in 2016. He is currently pursuing the Ph.D. degree in Computer Science with the Computer Science and Engineering department at the University of Texas at Arlington. His research interests include combinatorial optimization, machine learning, and data science.
\end{biographywithpic}

\begin{biographywithpic}
{Kyongsik Yun}{images/KyongsikYun.png} is a technologist at the Jet Propulsion Laboratory, California Institute of Technology, and a senior member of the American Institute of Aeronautics and Astronautics (AIAA). His research focuses on building brain-inspired technologies and systems, including deep learning computer vision, natural language processing, and multivariate time series models. He received the JPL Explorer Award (2019) for scientific and technical excellence in machine learning applications. Kyongsik received his B.S. and Ph.D. in Bioengineering from the Korea Advanced Institute of Science and Technology (KAIST).
\end{biographywithpic}

\begin{biographywithpic}
{Muhammad Adil}{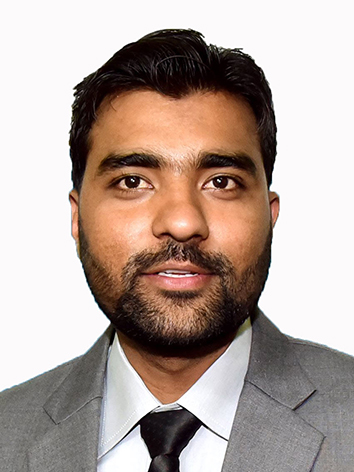} is currently working towards the Ph.D. degree in Electrical Engineering at University of Texas at Arlington. He received the B.S and M.S degrees in Electrical Engineering
from University of Engineering and Technology and Pakistan Institute of Engineering and Applied Sciences in 2012 and 2014. His research interests include developing numerical algorithms for solving large scale optimization problems.
\end{biographywithpic}

\begin{biographywithpic}
{Changrak Choi}{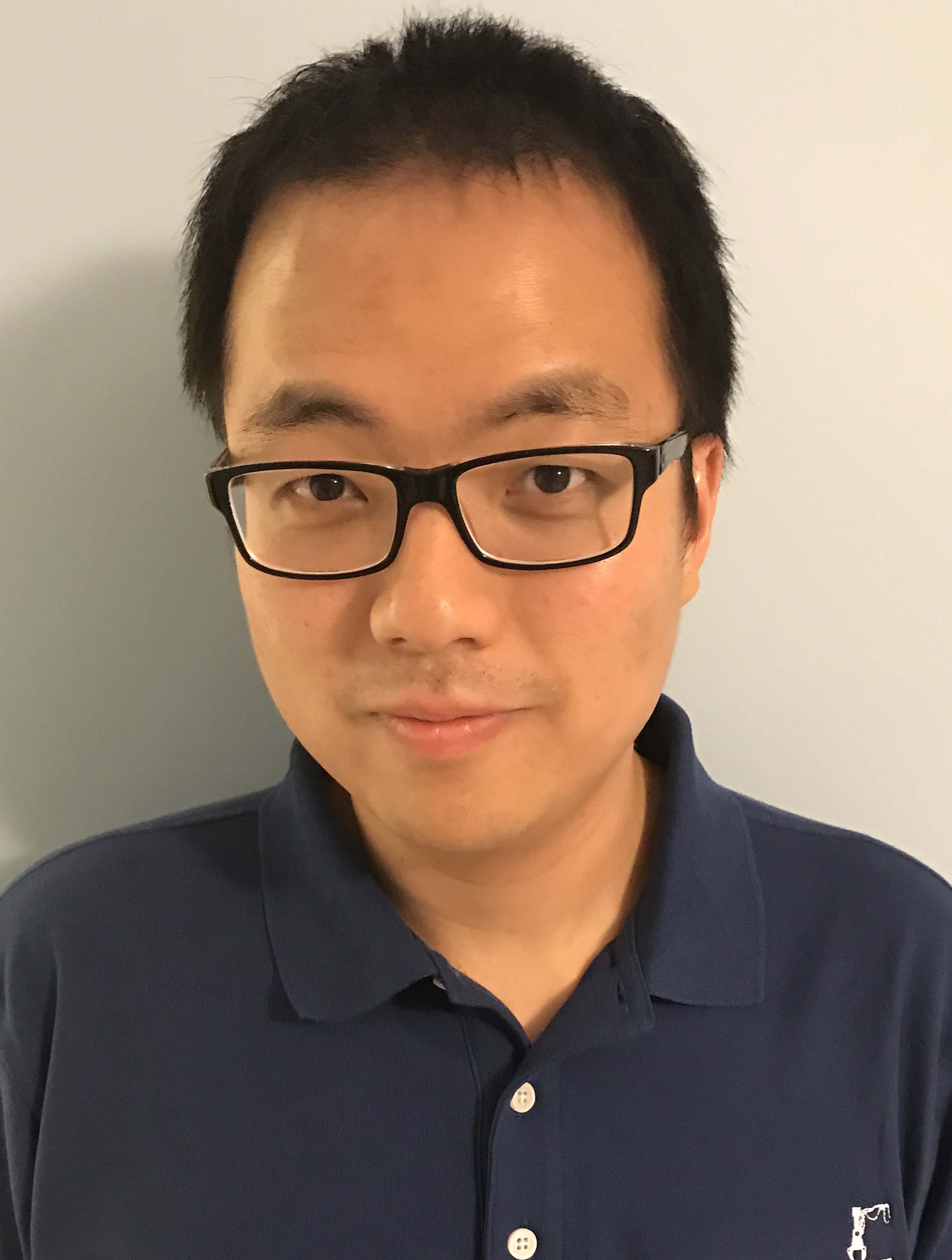}
is a Robotics Technologist in the Multi-Agent Autonomy group at JPL. His research focuses on trajectory and motion planning for autonomous systems that are dynamically interesting, drawing upon algorithms, optimization, and controls. The systems of interest range from ground to on-water and aerial vehicles as well as soft robots, with a special interest in multi-spacecraft systems. Changrak received his PhD from MIT as a member of Laboratory for Information and Decision Systems (LIDS) in Aerospace Robotics and Embedded Systems group. He earned MS and BS, both in Mechanical Engineering, from MIT and Seoul National University respectively.
\end{biographywithpic} 

\begin{biographywithpic}
{Ramtin Madani}{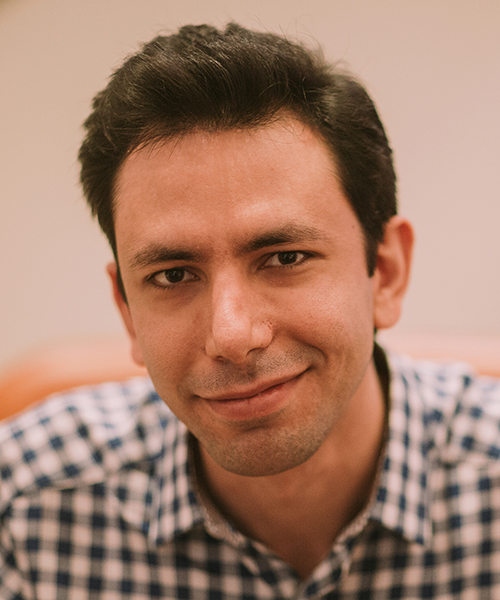} received the Ph.D. degree in electrical engineering from Columbia University, New York, NY, USA, in 2015. He was a Postdoctoral Scholar with the Department of Industrial Engineering and Operations Research at the University of California, Berkeley in 2016. He is an Assistant Professor with the Department of Electrical Engineering Department, University of Texas at Arlington, Arlington, TX, USA. His research interests include developing algorithms for optimization and control with applications in energy.
\end{biographywithpic}
\onecolumn
\appendix{}
\begin{figure}[h]
    \centering
	\includegraphics[scale=.35]{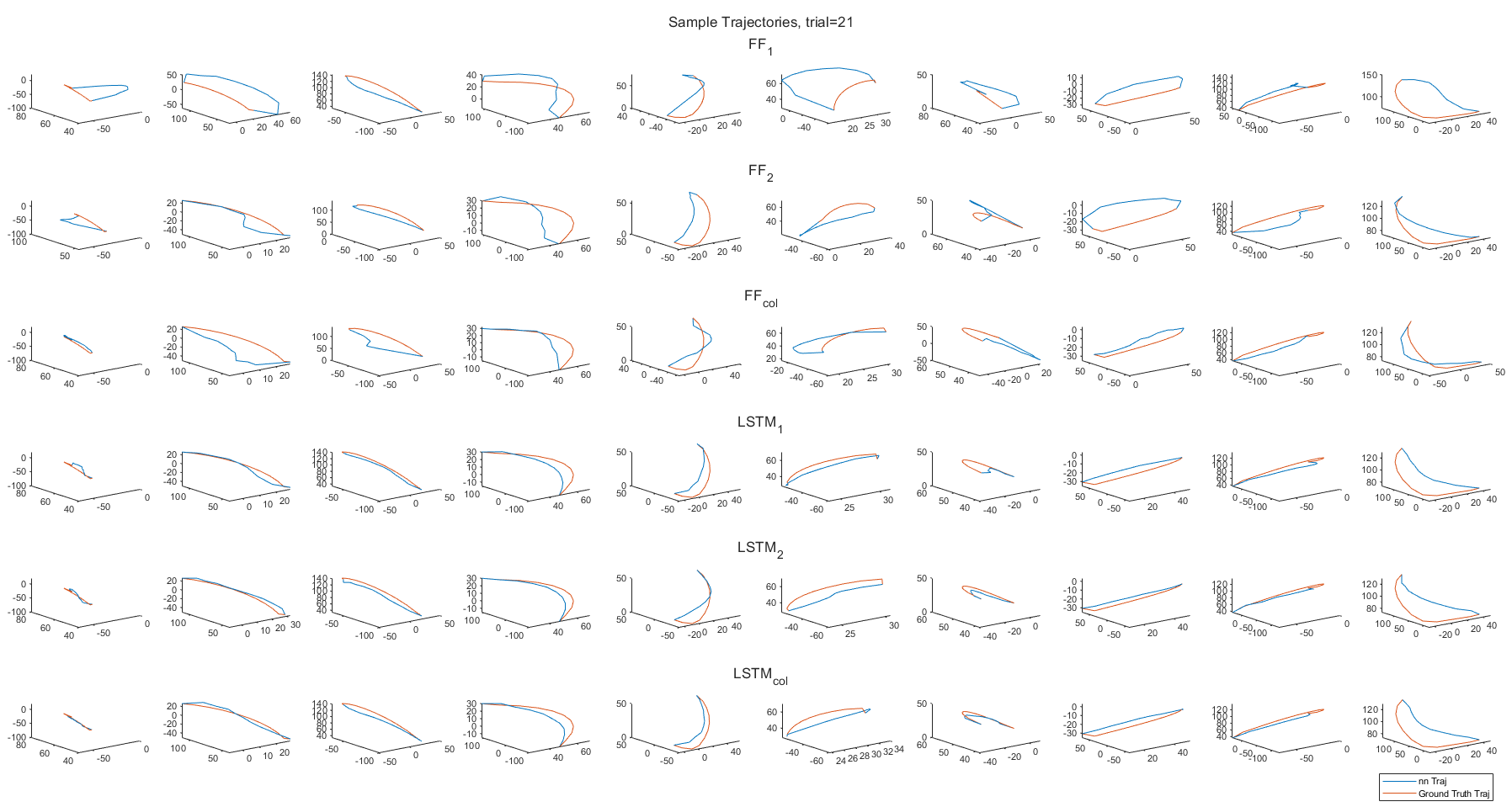}
	\caption{\textbf{Example of trajectory plans generated by each architecture for a sample 10 agent trial.} } \label{fig:exp1_traj}
\end{figure}
\end{document}